# Discovery of Generalizable TBI Phenotypes Using Multivariate Time-Series Clustering

Hamid Ghaderi [a, *], Brandon Foreman [b], Chandan K. Reddy [c], Vignesh Subbian [a, d]

[a] Department of Systems and Industrial Engineering, University of Arizona, Tucson, AZ, USA

[b] College of Medicine, University University of Cincinnati, Cincinnati, OH, USA

[c] Department of Computer Science, Virginia Tech, Arlington, VA, USA

[d] Department of Biomedical Engineering, University of Arizona, Tucson, AZ, USA

**Abstract**

Traumatic Brain Injury (TBI) presents a broad spectrum of clinical presentations and outcomes due to its inherent heterogeneity, leading to diverse recovery trajectories and varied therapeutic responses. While many studies have delved into TBI phenotyping for distinct patient populations, identifying TBI phenotypes that consistently generalize across various settings and populations remains a critical research gap. Our research addresses this by employing multivariate time-series clustering to unveil TBI's dynamic intricates. Utilizing a self-supervised learning-based approach to clustering multivariate time-Series data with missing values (SLAC-Time), we analyzed both the research-centric TRACK-TBI and the real-world MIMIC-IV datasets. Remarkably, the optimal hyperparameters of SLAC-Time and the ideal number of clusters remained consistent across these datasets, underscoring SLAC-Time's stability across heterogeneous datasets. Our analysis revealed three generalizable TBI phenotypes (α, β, and γ), each exhibiting distinct non-temporal features during emergency department visits, and temporal feature profiles throughout ICU stays. Specifically, phenotype α represents mild TBI with a remarkably consistent clinical presentation. In contrast, phenotype β signifies severe TBI with diverse clinical manifestations, and phenotype γ represents a moderate TBI profile in terms of severity and clinical diversity. Age is a significant determinant of TBI outcomes, with older cohorts recording higher mortality rates. Importantly, while certain features varied by age, the core characteristics of TBI manifestations tied to each phenotype remain consistent across diverse populations.

## 1. Introduction

Traumatic Brain Injury (TBI) is a prevalent global health concern affecting a range of populations, from young athletes with sports-related injuries to military personnel exposed to combat-related trauma, and older adults prone to falls [1–4]. Every year, millions experience the consequences of TBI, with outcomes spanning mild concussions to severe cognitive dysfunctions [1,5]. The diverse range of clinical presentations underscores the inherent heterogeneity of TBI [6–8]. This heterogeneity complicates prognosis, treatment, and research. TBI patients with apparently similar injuries can have varied recovery patterns and therapeutic responses. This variability suggests different TBI phenotypes, each with its own trajectory and treatment needs. By categorizing TBI into specific subgroups, clinicians provide more precise interventions, while also aiding researchers in designing more targeted studies [9–11]. Such an individualized approach can potentially improve recovery outcomes, minimize long-term complications, and enhance quality of life for those affected by TBI. Moreover, personalized treatment for TBI can mitigate treatment costs by targeting specific patient needs, thereby minimizing unnecessary procedures and expediting recovery [12].

---

[*] Corresponding author: ghaderi@arizona.edu

In recent years, clustering analysis has been increasingly used for subgrouping TBI patients and identifying TBI phenotypes [13]. However, these studies often rely on non-temporal data, which only reflects a snapshot of the injury and fails to capture the evolving characteristics of TBI, crucial elements for comprehensively understanding the condition and developing effective treatments [14]. This highlights the need for methodologies that incorporate time-series data, thereby more accurately addressing the temporal progression of TBI in phenotyping. The existing methodologies for clustering time-series data are only effective when there are no missing values, which is rarely the case for TBI data [14,15]. Common techniques such as data imputation or interpolation are routinely used for addressing missing values in time-series data in various applications [16]. However, in the context of TBI, where each clinical variable can have profound implications, applying these techniques without thorough knowledge can lead to significant biases. For instance, imputing average values for missing intracranial pressure readings could hide important details needed for the diagnosis and prevention of secondary brain injury. Similarly, interpolating cognitive function scores without considering the unique recovery trajectory of each TBI patient could lead to misinterpretation of their recovery progress, potentially resulting in inappropriate clinical interventions. These scenarios highlight the need for handling of missing values in TBI data without resorting to common data imputation or interpolation methods to ensure that clustering analyses yield valid and actionable insights for patient care.

Another major gap in the existing studies is lack of external validation of the identified TBI phenotypes. External validation of TBI phenotypes is a critical step in ensuring their broader applicability and reliability. By testing phenotypes in diverse patient populations, often across different geographical and healthcare settings, external validation confirms that the findings are not just relevant to the specific cohort in which they were initially identified. This process helps to establish the generalizability of the phenotypes, ensuring that they hold true across various demographic, clinical, and environmental contexts. Such validation is essential in TBI phenotyping due to the highly variable nature of the condition, which can be influenced by factors like injury severity, patient age, and co-existing health issues. Despite this need, the existing works have mainly studied only one dataset for phenotype identification, overlooking the necessity to corroborate these findings across other datasets. This oversight in validation could lead to the implementation of phenotype-specific clinical interventions that are not universally applicable, risking suboptimal or even harmful patient care. This research gap highlights the need for a methodological approach that not only identifies but also validates TBI phenotypes with high reliability and applicability across various patient cohorts.

To address the gaps, in this study, we use a self-supervised learning-based approach for clustering multivariate time-series data with missing values (SLAC-Time) for identifying TBI phenotypes. Unlike traditional clustering approaches that treat each multivariate time-series data as a matrix with specific dimensions, SLAC-Time's transformer-based architecture treats each input of multivariate time-series data as a set of observation triplets, enabling us to cluster the data without resorting to any data imputation or interpolation methods [14]. SLAC-Time leverages self-supervision to benefit from strong representation learning capabilities of transformer models in clustering. Specifically, it utilizes representations learned from input data for clustering, thereby mitigating the impact of noise and outliers in the raw input. To discover generalizable TBI phenotypes, we perform clustering analysis on two different TBI datasets: the Transforming Research and Clinical Knowledge in Traumatic Brain Injury (TRACK-TBI) dataset [17] and the Medical Information Mart for Intensive Care (MIMIC)-IV dataset [18,19]. Notably, we use a feature set for clustering that is common to both TRACK-TBI and MIMIC-IV datasets. Our selection of these two datasets is deliberate, given their inherent differences. TRACK-TBI is primarily a multicenter research dataset, with the intent of advancing TBI research. It includes detailed clinical data along with long-term outcomes, such as the Extended Glasgow Outcome Scale (GOSE) score, providing a comprehensive view of TBI progression and recovery [20]. On the other hand, MIMIC-IV is a real-world clinical dataset from a single institution. It includes routinely collected clinical data in electronic health records, but without long-term outcome data that are inherent to research datasets such as TRACK-TBI. By comparing the clustering findings from the TRACK-TBI research data to those of MIMIC-IV, we aim to identify phenotypes that are consistent across diverse patient populations and settings. We anticipate that the resulting phenotypes will be both insightful and broadly applicable for developing tailored clinical treatments. Figure 1 demonstrates a high-level overview of the work.

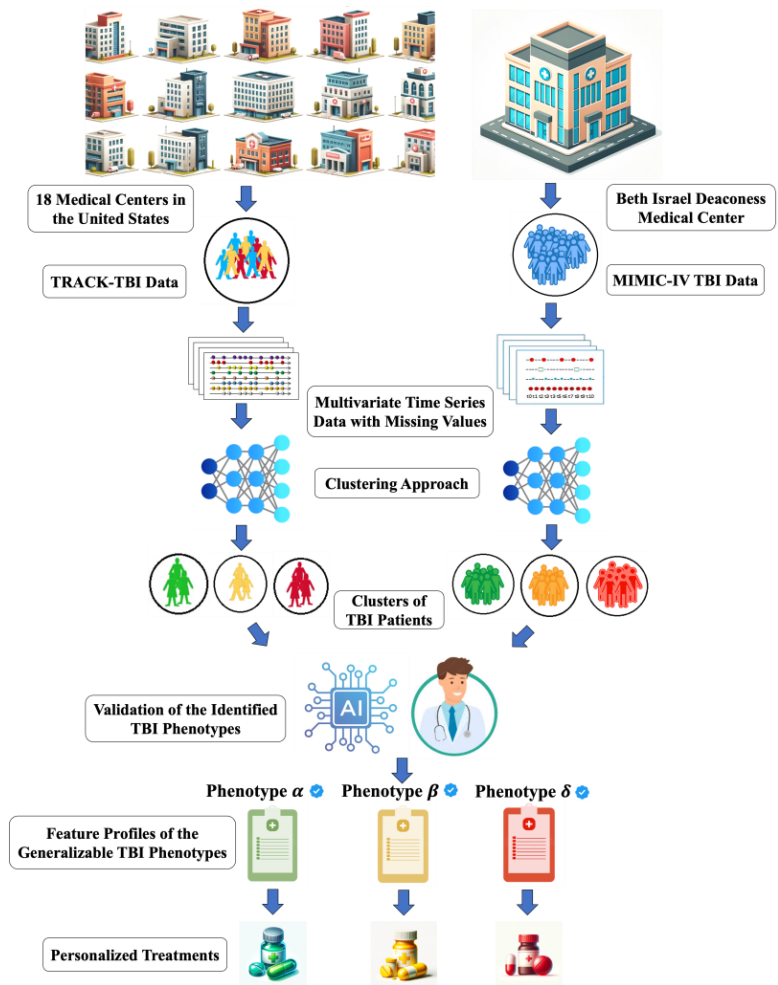

Figure 1. A high-level overview of methodology in identifying generalizable TBI phenotypes.

The following are contribution of this work:

- This study transcends the traditional static TBI phenotyping methods by incorporating the complexity of TBI progression through multivariate time-series data analysis. This allows for the identification of TBI phenotypes that reflect the evolving pathophysiology of the injury over time.
- By leveraging the complementary strengths of the TRACK-TBI and MIMIC-IV datasets, this study provides a robust validation of these phenotypes across varied clinical and research settings, enhancing the generalizability and applicability of our findings.
- This study employs the SLAC-Time algorithm, which does not rely on data imputation for handling missing values. This allows us to preserve the integrity of the original datasets, ensuring that our findings are robust and devoid of biases typically introduced by conventional imputation techniques. Furthermore, by conducting clustering on the representations learned from the transformer model, the method effectively reduces noise and mitigates the impact of outliers in the raw input data.

The rest of this paper is organized as follows. Section 2 provides the related work on phenotyping TBI patients using clustering analysis. Section 3 presents the materials and methods, describing the datasets used, data preprocessing steps, details of the SLAC-Time clustering algorithm, and TBI phenotype evaluation and validation methodologies. The results from the clustering analysis on the TRACK-TBI and MIMIC-IV datasets are reported in Section 4, including a characterization of the identified TBI phenotypes, comparative analysis, and evaluation of reproducibility. We discuss key implications and limitations of our findings in Section 5. The paper concludes with Section 6, which encapsulates our study's results and suggests avenues for future research.

## 2. Related Work

The field of TBI phenotyping has seen a variety of approaches, with several studies leveraging clustering algorithms to stratify patients. This section synthesizes the key trends, methodologies, and gaps in current research, providing a holistic view of the field.

Maddux et al. [21] applied K-medoids clustering to insurance claims data spanning the first year post-discharge in pediatric TBI survivors. Their study stands out for its comprehensive inclusion of diverse variables such as hospital readmissions, emergency and outpatient visits, and therapy sessions. They identified four functional outcome TBI phenotypes, varying from minimal to those with extensive post-discharge healthcare needs and morbidities. Adding a new dimension to TBI phenotyping, Fujiwara et al. [22] focused on coagulation phenotypes. By applying K-means clustering to acute-phase coagulation markers, they identified five TBI phenotypes, highlighting the link between coagulation, skull fractures, and trauma severity, thus offering a unique perspective in TBI research.

Yeboah et al. [23] used ensemble clustering model based on baseline measurements to identify TBI phenotypes. They identified six phenotypes of TBI patients, differentiated by their injury mechanisms, including motor vehicle accidents, diverse incidents like pedestrian accidents or sports, falls from moving or stationary objects, accidents involving motorcycles or all-terrain vehicles, and assaults, each characterized by distinct initial severities and recovery trajectories. Focusing on mild TBI, Si et al. [24] employed sparse hierarchical clustering on a set of 12 clinical variables. This study is particularly significant for its contribution to understanding mild TBI, as it identified five distinct phenotypes, each associated with different clinical outcomes and prevalence rates.

Folweiler et al. [9] utilized a wrapper framework with generalized low-rank models to select the relevant clinical features in a TBI dataset, which were then used to cluster patients into phenotypes using a partitioning around medoids (PAM) clustering algorithm. This approach identified three TBI phenotypes with unique clinical feature profiles and long-term functional outcomes, including one characterized by mild anemia, another with normal hematological values but lower platelet count and elevated prothrombin time, and a third presenting a severe clinical profile with signs of thrombocytopenia, anemia, and coagulopathy. The study further validated these phenotypes in an external dataset using a K-nearest neighbors algorithm, suggesting that these phenotypes could be generalizable across a broad range of TBI severity.

Addressing the challenge of missing data in TBI phenotyping, Akerlund et al. [25] developed a novel approach using unsupervised learning techniques rooted in probabilistic graph models. Their methodology, which did not rely on imputing missing feature values, revealed six TBI phenotypes, including a spectrum from mild TBI in older patients often on anticoagulants to severe TBI in younger patients with significant metabolic and respiratory acidosis. Complementing this, Ghaderi et al. [14] developed SLAC-Time, an innovative method for clustering multivariate time-series data with missing values without resorting to data imputation or aggregation methods, and applied it to a cohort of TBI patients characterized by a combination of static and dynamic clinical variables. This approach successfully identified three distinct TBI phenotypes, each differentiated by unique clinical variables and associated outcome endpoints.

Despite these advancements, a common limitation in the field is the focus on non-temporal data and a lack of external validation of identified phenotypes. Our research aims to bridge these gaps by integrating both temporal and non-temporal variables in identifying TBI phenotypes and validating these phenotypes with an independent dataset. This approach not only enhances the generalizability of our findings but also contributes to a more comprehensive understanding of TBI phenotypes, potentially leading to more tailored and effective treatment strategies.

## 3. Materials and Methods

3.1. Datasets

*TRACK-TBI Dataset*: The TRACK-TBI dataset includes 2996 TBI patients, collected across 18 medical centers in the United States [17]. It comprises around 250 variables, including patient demographics, clinical assessments, radiological findings, and outcomes. For the purposes of our study, we only included those TBI patients for whom both temporal and non-temporal data were available. As a result, a total of 2,160 TBI patients from the TRACK-TBI dataset were included in our analysis.

*MIMIC-IV Dataset*: The MIMIC-IV dataset is a publicly accessible resource, covering over 60,000 ICU admissions at the Beth Israel Deaconess Medical Center between 2008 and 2019 [18,19]. This dataset is structured as a relational database and includes a wide array of both temporal and non-temporal data. The data categories include patient demographics, vital signs, laboratory test results, medications administered, and clinical outcomes. Our study included 3,140 TBI patients from this extensive dataset for analysis.

3.2. Data Preprocessing

*Variable Selection*: We selected non-temporal and time-series variables common to both the TRACK-TBI and MIMIC-IV datasets, discarding all others. Our analysis incorporated 20 non-temporal and 38 time-series variables. In addition, two outcome variables—ICU length of stay and mortality rate—were included to evaluate the TBI phenotypes identified in both datasets. Table 1 provides the summary statistics of the key features of the TBI patients in TRACK-TBI and MIMIC-IV datasets.

Table 1. Summary statistics of key features for TBI patients in the TRACK-TBI and MIMIC-IV datasets

| Features | TRACK-TBI | MIMIC-IV |
|---|---|---|
| Demographics | | |
| Total subjects, n | 2160 | 3140 |
| Age, mean ± SD | 40 ± 18 | 61 ± 22 |
| Age≤30, n (%) | 874 (41%) | 443 (14%) |
| 30<Age≤45, n (%) | 481 (22%) | 346 (11%) |
| 45<Age≤60, n (%) | 455 (21%) | 594 (19%) |
| Age>60, n (%) | 350 (16%) | 1757 (56%) |
| Sex (Female, n (%)) | 648 (30%) | 1170 (37%) |
| ED Examination | | |
| ED Glucose, mean ± SD | 139 ± 56 | 135 ± 47 |
| ED Hemoglobin, mean ± SD | 13.9 ± 1.8 | 12.4 ± 1.8 |
| ED INR, mean ± SD | 1.10 ± 0.36 | 1.27 ± 0.17 |
| ED White Blood Cell, mean ± SD | 12.4 ± 5.45 | 12 ± 6.13 |

*Handling Missing Data*: For handling missing data in non-temporal variables, an iterative imputation technique was employed. Meanwhile, missing values in time-series variables were managed through SLAC-Time, avoiding the need for imputation.

*Removing Outliers*: To enhance data quality and maintain clinical relevance, we imposed specific clinical criteria to set numerical ranges for each variable, effectively eliminating outliers. In defining the ranges for the clinical variables, we established the highest and lowest possible values that could be realistically experienced by both healthy individuals and TBI patients. This was achieved through a rigorous process that involved a comprehensive review of existing medical literature and clinical guidelines to identify physiological limits for each variable. This method provided a broader and more clinically relevant range compared to the ranges defined by traditional outlier removal approaches, which often rely on statistical distributions that may not accurately reflect TBI realities.

*Data Normalization and Encoding*: We normalized both time-series and numerical non-temporal variables using the Z-score normalization method to have zero mean and unit variance. Additionally, we employed one-hot encoding to create a binary vector for each category within the categorical variables.

*Time-Series Standardization*: Due to varying lengths of time-series data among patients, we standardized these to a five-day duration corresponding to the initial ICU admission period. Within each time-series variable, average values were calculated at hourly intervals, resulting in 120 time steps per variable.

3.3. Clustering Algorithm

We utilize SLAC-Time, built on a self-supervised Transformer model known as STraTS [26], for representation learning of $N$ unlabeled samples represented by $\mathcal{D} = \{(d^k, T^k)\}_{k=1}^{N}$ where the $k^{th}$ sample includes a non-temporal vector $d^k \in \mathbb{R}^D$ and multivariate time-series data $T^k$. STraTS maps the data into a fixed-dimensional vector space $\mathbb{R}^d$. Each multivariate time-series data $T^k$ of length $n$ is represented as a set of $n$ observation triplets $T^k = \{(t_i, f_i, v_i)\}_{i=1}^{n}$ where $t_i \in \mathbb{R}_{\geq 0}$ is the time, $f_i$ is the feature name, and $v_i \in \mathbb{R}$ is the value. The initial embedding $e_i$ for a triplet is obtained by summing the feature, value, and time embeddings, denoted as $e_i^f$, $e_i^v$, and $e_i^t$ respectively:

$$e_i = e_i^f + e_i^v + e_i^t \in \mathbb{R}^d \tag{1}$$

where the feature embedding $e_i^f$ is obtained from a basic lookup table, while the value embedding $e_i^v$ and time embedding $e_i^t$ are computed through one-to-many Feed-Forward Networks (FFNs). The embedding for non-temporal variables is also acquired by processing $d$ through an FNN.

SLAC-Time is organized into three modules: self-supervision, pseudo-label extraction, and classification [14].

(1) Self-Supervision Module: STraTS is pretrained via a forecasting task. The forecasting task output is generated by feeding the concatenated embeddings of both no-temporal and time series variables through the following layer.

$$\tilde{z} = W_s[e^d \ e^T] + b_s \in \mathbb{R}^{|\mathcal{F}|} \tag{2}$$

In this task, to handle missing values in the forecasting outputs, we employ a masked Mean Squared Error (MSE) loss, defined as:

$$\mathcal{L}_{ss} = \frac{1}{|N'|} \sum_{k=1}^{N'} \sum_{j=1}^{|\mathcal{F}|} m_j^k (\tilde{z}_j^k - z_j^k)^2 \tag{3}$$

Here, $N' \geq N$ is the number of samples, $F$ is the feature set, $m_j^k$ is the forecast mask, and $\tilde{z}_j^k$ and $z_j^k$ are the predicted and actual values, respectively.

(2) Pseudo-label Extraction Module: After pre-training, we use the embeddings $f_\theta((d^n, T^n))$ for K-means clustering to obtain pseudo-labels $y_n$, by solving:

$$\min_{C \in \mathbb{R}^{d \times k}} \frac{1}{N} \sum_{n=1}^{N} \min_{y_n \in \{0,1\}^k} \|f_\theta((d^n, T^n)) - C y_n\|_2^2 \text{ such that } y_n^\top 1_k = 1 \tag{4}$$

Here, $C$ is the centroid matrix, $N$ is the number of subjects, and $k$ is the number of clusters.

(3) Classification Module: A classifier $g_W$ is trained using the pseudo-labels $y_n$ and the representations $f_\theta((d^n, T^n))$ to minimize the loss function $\ell$:

$$\min_{\theta, W} \frac{1}{N} \sum_{n=1}^{N} \ell\left(g_W\left(f_\theta((d^n, T^n))\right), y_n\right) \tag{5}$$

Here, $\theta$ and $W$ are the parameters of STraTS and the classifier, respectively.

SLAC-Time operates in an iterative manner, alternating between pseudo-label extraction and classifier parameter updates to enhance clustering and classification performance.

2.4. Implementation and Training Details

For the initial representation learning of the multivariate time-series data, we employed time-series forecasting as a proxy task to pre-train our model. We delineated our observation windows as {24, 48, 72, 96, 118} hours and established a subsequent 2-hour period as the prediction window. Importantly, our analysis only considered records that contained at least one time-series data entry within both the observation and prediction windows. The dataset utilized for time-series forecasting was partitioned into training and validation sets at an 80:20 ratio. Concurrently, the target task of SLAC-Time is to subgroup TBI patients based on the variables in the dataset. Similar to the forecasting task, for the target task, these TBI patients were allocated into training and validation subsets following an 80:20 distribution.

We developed our SLAC-Time model using Keras, backed by TensorFlow. We trained both the proxy and target task models with a batch size of 8, using the Adam optimizer. For the proxy task, training was stopped if the validation loss did not decrease for ten consecutive epochs. The target task training proceeded for 500 iterations, and each iteration consisted of 200 epochs. As with the proxy task, training for each iteration was stopped if the validation loss did not decrease for ten consecutive epochs. We performed all computational experiments on an NVIDIA Tesla P100 GPU.

2.5. Clustering Evaluation

We utilized three clustering evaluation metrics to guide our hyperparameter selection: the Silhouette Score [27], the Calinski-Harabasz Score [28], and the Davies-Bouldin Score [29]. The Silhouette Score, measuring intra-cluster similarity against neighboring clusters and aiming for values close to 1, is useful for assessing the clarity of cluster separation. The Calinski-Harabasz Score, quantifying the ratio of between-cluster variance to within-cluster variance and favoring higher values, helps evaluate cluster compactness and separation. Lastly, the Davies-Bouldin Score, which calculates the average similarity ratio between each cluster and its nearest counterpart, is preferred to be lower, indicating well-separated clusters with minimal within-cluster dispersion.

2.6. Fine-Tuning Hyperparameters

The performance of the SLAC-Time clustering approach largely depends on its hyperparameters, including the number of transformer blocks (M), the number of layers (d), the number of attention heads (h), and the number of clusters (K). We systematically varied these hyperparameters for evaluation on the TRACK-TBI and MIMIC-IV datasets. For the TRACK-TBI dataset, the configuration with M=1, d=8, h=2, and K=3 was superior across all metrics, resulting in a Silhouette Score of 0.33, a Calinski-Harabasz Score of 729.12, and a Davies-Bouldin Score of 1.36 (Table 2). This configuration also resulted in the best metrics for the MIMIC-IV dataset, showing scores of 0.25, 884.95, and 1.47, respectively (Table 3). The same optimal architecture for both datasets underscores the robustness of SLAC-Time clustering approach. Moreover, the choice of K=3 as the optimal number of clusters in both datasets shows the existence of three potential generalizable TBI phenotypes.

Table 2. Clustering evaluation scores for different hyperparameter configurations on the TRACK-TBI dataset. (SS: Silhouette Score; CHS: Calinski-Harabasz Score; DBS: Davies-Bouldin Score). The configuration M=1, d=8, h=2, and K=3 achieved the best performance across all metrics, suggesting it as the optimal setup for the TRACK-TBI dataset.

| Hyperparameters | K=3 | | | K=4 | | | K=5 | | |
| --- | --- | --- | --- | --- | --- | --- | --- | --- | --- |
| | SS | CHS | DBS | SS | CHS | DBS | SS | CHS | DBS |
| M=1, d=8, h=2 | **0.33** | **729.12** | **1.36** | 0.25 | 487.60 | 1.52 | 0.23 | 704.98 | 1.48 |
| M=1, d=8, h=4 | 0.21 | 713.20 | 1.75 | 0.24 | 416.43 | 1.57 | 0.12 | 380.90 | 1.94 |
| M=1, d=16, h=2 | 0.15 | 232.99 | 2.34 | 0.07 | 251.96 | 2.51 | 0.17 | 320.07 | 2.14 |
| M=1, d=16, h=4 | 0.10 | 318.56 | 2.78 | 0.10 | 231.88 | 2.33 | 0.12 | 248.76 | 2.20 |
| M=1, d=32, h=4 | 0.06 | 147.41 | 3.20 | 0.05 | 114.07 | 3.67 | 0.04 | 105.24 | 3.19 |
| M=1, d=32, h=8 | 0.07 | 159.41 | 3.26 | 0.05 | 125.36 | 3.13 | 0.05 | 98.13 | 3.27 |
| M=2, d=8, h=2 | 0.17 | 397.07 | 1.98 | 0.18 | 394.96 | 1.95 | 0.16 | 507.74 | 1.75 |
| M=2, d=8, h=4 | 0.21 | 691.70 | 1.65 | 0.15 | 315.71 | 1.82 | 0.12 | 281.04 | 1.88 |
| M=2, d=16, h=2 | 0.13 | 268.70 | 2.63 | 0.08 | 234.97 | 2.55 | 0.11 | 206.08 | 2.56 |
| M=2, d=16, h=4 | 0.10 | 218.56 | 2.53 | 0.12 | 235.63 | 2.55 | 0.09 | 205.47 | 2.33 |
| M=2, d=32, h=2 | 0.08 | 174.17 | 2.96 | 0.06 | 134.34 | 2.96 | 0.04 | 92.03 | 3.32 |
| M=2, d=32, h=4 | 0.08 | 183.93 | 2.93 | 0.05 | 112.80 | 3.17 | 0.04 | 92.66 | 3.59 |
| M=2, d=64, h=4 | 0.06 | 139.11 | 3.25 | 0.06 | 120.44 | 3.14 | 0.06 | 124.47 | 2.91 |
| M=2, d=128, h=4 | 0.06 | 136.58 | 3.35 | 0.05 | 109.47 | 3.44 | 0.05 | 92.52 | 3.20 |

Table 3. Clustering evaluation scores for different hyperparameter configurations on the MIMIC-IV dataset. (SS: Silhouette Score; CHS: Calinski-Harabasz Score; DBS: Davies-Bouldin Score). The configuration M=1, d=8, h=2, and K=3 achieved the best performance across all metrics, suggesting it as the optimal setup for the MIMIC-IV dataset.

| Hyperparameters | K=3 | | | K=4 | | | K=5 | | |
| --- | --- | --- | --- | --- | --- | --- | --- | --- | --- |
| | SS | CHS | DBS | SS | CHS | DBS | SS | CHS | DBS |
| M=1, d=8, h=2 | **0.25** | **884.95** | **1.47** | 0.24 | 680.51 | 1.67 | 0.13 | 458.59 | 1.89 |
| M=1, d=8, h=4 | 0.23 | 631.74 | 1.78 | 0.13 | 707.22 | 1.87 | 0.16 | 671.80 | 1.72 |
| M=1, d=16, h=2 | 0.17 | 457.76 | 2.02 | 0.13 | 547.68 | 2.43 | 0.14 | 463.17 | 2.26 |
| M=1, d=16, h=4 | 0.13 | 455.67 | 2.41 | 0.10 | 460.91 | 2.41 | 0.10 | 478.54 | 2.20 |
| M=1, d=32, h=4 | 0.08 | 280.06 | 2.81 | 0.11 | 269.57 | 2.99 | 0.09 | 257.99 | 2.85 |
| M=1, d=32, h=8 | 0.05 | 228.10 | 3.36 | 0.06 | 197.85 | 3.12 | 0.05 | 180.37 | 3.15 |
| M=2, d=8, h=2 | 0.14 | 810.25 | 2.06 | 0.16 | 609.06 | 1.89 | 0.16 | 669.63 | 1.77 |
| M=2, d=8, h=4 | 0.20 | 680.40 | 1.87 | 0.12 | 691.99 | 1.85 | 0.22 | 799.93 | 1.55 |
| M=2, d=16, h=2 | 0.09 | 342.08 | 2.65 | 0.09 | 332.93 | 2.40 | 0.13 | 449.23 | 2.20 |
| M=2, d=16, h=4 | 0.13 | 463.70 | 2.19 | 0.09 | 296.01 | 2.44 | 0.09 | 274.68 | 2.41 |
| M=2, d=32, h=2 | 0.11 | 358.12 | 2.69 | 0.07 | 246.35 | 2.92 | 0.06 | 162.41 | 3.38 |
| M=2, d=32, h=4 | 0.10 | 311.41 | 2.82 | 0.07 | 237.12 | 2.94 | 0.06 | 196.55 | 3.34 |
| M=2, d=64, h=4 | 0.13 | 438.68 | 2.47 | 0.12 | 338.72 | 2.91 | 0.12 | 329.10 | 2.51 |
| M=2, d=128, h=4 | 0.14 | 416.29 | 2.85 | 0.14 | 368.75 | 2.32 | 0.14 | 346.96 | 2.34 |

2.7. Characterization and Comparative Analysis of TBI Phenotypes

To understand the unique characteristics of identified TBI phenotypes, we analyzed variables previously shown to significantly influence GOSE score of TBI patients. In this regard, we focused on clinical variables, including age, sex, Glasgow Coma Scale (GCS) motor score, GCS eye score, glucose, hemoglobin, white blood cell count (WBC), hematocrit, International Normalized Ratio (INR), and Activated Partial Thromboplastin Time (aPTT). For comparative analyses, non-temporal variables were represented numerically using means and standard deviations, or as counts and proportions (Table 4). On the other hand, time-series variables were described using phenotype-specific averages, accompanied by 95% confidence intervals (Figures 3-9). We performed univariate analyses on both non-temporal and time-series variables to identify potential differences between the phenotypes. The Kruskal-Wallis test was used to determine the statistical significance of observed differences. Differences between phenotypes were considered significant if the associated p-values were below 0.05. To validate the consistency of TBI phenotypes across datasets, we conducted a comparative study of the phenotypes identified in each dataset.

2.8. External Validation Methodology

To assess the reproducibility of TBI phenotypes identified in the TRACK-TBI dataset via the SLAC-Time clustering, we utilized a transformer-based classifier, informed by three phenotype labels and the same feature set used in the clustering analysis. Specifically, TRACK-TBI subjects were split into two: 85% for training/validation and 15% for testing. This training/validation set was further subjected to 10-fold stratified cross-validation, ensuring each fold maintained a balanced phenotype representation, critical due to the phenotype heterogeneity.

The process of each fold included:

- Model Initialization: A transformer-based classifier was configured in line with the optimal setup of the SLAC-Time model.
- Transfer Learning: We trained the classifier with the weights obtained from the clustering task, expediting convergence and possibly enhancing performance.
- Training: With Adam optimizer and sparse categorical cross-entropy loss monitoring, the model was trained using the current fold, leveraging the validation fold for early stopping.
- Model Evaluation: Post-training performance was assessed on the 15% test set.

Following training, the classifier was applied to the MIMIC-IV dataset. Then, a cross-match permutation test based on Euclidean distance was used to compare the predicted labels with the TRACK-TBI phenotypes. We also employed the same methodology to assess the reproducibility of MIMIC-IV phenotypes in the TRACK-TBI dataset.

4. Results

Our study identified three distinct TBI phenotypes in the TRACK-TBI and MIMIC-IV datasets, each with unique characteristics and clinical outcomes:

- Phenotype α shows the lowest mortality rates and shortest ICU stays. It includes the youngest group of patients with better neurological outcomes, moderate metabolic stress, optimal hydration and oxygenation, the mildest inflammatory response, and a more efficient coagulation profile.
- Phenotype β exhibits the highest mortality rates and longest ICU stays, and is characterized by the most severe neurological impairments, elevated metabolic stress, the lowest levels of hydration and oxygenation, the strongest inflammatory response, and a compromised coagulation profile.
- Phenotype γ, consisting of the oldest patient group, has clinical features that are intermediate in severity. This phenotype shares more similarities with phenotype α, exhibiting moderately elevated metabolic stress, balanced hydration and oxygenation, moderate inflammatory response, and a relatively balanced coagulation profile.

Subsequent sections (4.1.1 to 4.1.4) provide a detailed comparative analysis of demographics, non-temporal and time-series clinical markers, and outcomes for these phenotypes across the two datasets.

4.1. Comparison of Demographics and Non-Temporal Clinical Features of TBI Phenotypes

*Age Distribution*: The age distribution across TBI phenotypes shows phenotype α as consistently the youngest group and phenotype γ the oldest in both TRACK-TBI and MIMIC-IV datasets, as detailed in Table 4. Phenotype α's average age is $38 \pm 18$ years in TRACK-TBI, increasing to $59 \pm 22$ years in MIMIC-IV, while phenotype γ averages at $43 \pm 18$ years and $69 \pm 20$ years in the respective datasets. This age pattern highlights the reliability of TBI phenotypes across varied patient demographics.

*Sex Distribution*: In the TRACK-TBI dataset, male representation is notable across all phenotypes, with 68%, 74%, and 76% for Phenotypes α, β, and γ respectively. Similarly, in the MIMIC-IV dataset, males constitute 63% of Phenotype α and approximately 63% for both Phenotypes β and γ. Although there is a slight reduction in the male-to-female ratio in the MIMIC-IV dataset relative to TRACK-TBI, both datasets consistently demonstrate that male patients are predominantly represented in each TBI phenotype, underscoring that male individuals may be more susceptible to certain TBI manifestations.

*Non-Temporal Clinical Variables in Emergency Departments:* Clinical variables recorded upon TBI patients' arrival at the emergency department provide insights into the immediate physiological reactions post-injury. The TRACK-TBI dataset reveals distinct TBI phenotypes ($p < 0.05$), with a range of metabolic and coagulation responses. Phenotype α in this dataset displays a nearly balanced metabolic response, indicated by a glucose level of $130 \pm 50$ mg/dL, and a robust oxygen-carrying capacity (hemoglobin level of $14.3 \pm 1.6$ g/dL). In contrast, phenotype β exhibits elevated metabolic stress (glucose level of $171 \pm 65$ mg/dL) and a slight reduction in oxygen transport (hemoglobin level of $13.3 \pm 1.8$ g/dL). Phenotype γ presents intermediate values between phenotypes α and β. Furthermore, in terms of coagulation and inflammatory responses, phenotype α shows the lowest mean INR value ($1.07 \pm 0.2$) and a moderate WBC count ($11.6 \pm 4.7$), indicating a relatively normal coagulation status and a mild inflammatory response. Phenotype β, with the highest mean INR ($1.24 \pm 0.8$) and WBC count ($15.5 \pm 6.4$), suggests a greater propensity for bleeding complications and a pronounced inflammatory or immune response, indicative of more severe trauma. Phenotype γ, with its slightly elevated INR ($1.10 \pm 0.2$) and higher WBC count ($13.7 \pm 6.6$) compared to phenotype α, occupies an intermediate position, reflecting moderate severity.

In comparison, the MIMIC-IV dataset, influenced by an older patient demographic, presents a parallel yet distinct TBI phenotypes ($p < 0.05$). Phenotype α in MIMIC-IV, while exhibiting similar glucose levels ($133 \pm 53$ mg/dL) to its TRACK-TBI counterpart, shows a decreased oxygen-carrying capacity (hemoglobin level of $12.5 \pm 1.7$ g/dL). This reduction might be attributed to age-related hematopoietic changes or comorbidities. Phenotype β in MIMIC-IV continues this trend of decreased hemoglobin levels ($11.4 \pm 1.9$ g/dL) alongside a slight rise in glucose ($139 \pm$

39 mg/dL). Phenotype γ falls between phenotypes α and β, reflecting moderate severity of TBI. Regarding coagulation and inflammation, phenotype α in MIMIC-IV, with an INR value of 1.25 ± 0.1 and a WBC count of 11.7 ± 5.7, suggests a mild coagulation abnormality and a similar level of inflammation or immune response as seen in TRACK-TBI. Phenotype β, marked by the highest INR value in the dataset (1.30 ± 0.1), indicates a significant risk of coagulopathy, potentially reflective of more severe brain injury. Furthermore, it has the highest WBC count (12.9 ± 7.7), suggesting an elevated inflammatory response. Phenotype γ, with an INR of 1.26 ± 0.1 and a WBC count of 11.9 ± 4.7, presents a moderate severity of TBI, with a balance between coagulation risks and inflammatory response.

The comparative analysis of non-temporal variables demonstrates consistent results across both datasets. Phenotype α typically presents milder clinical indicators, phenotype β manifests increased physiological alterations, and phenotype γ sits intermediate. However, this static snapshot does not capture the evolving nature of TBI phenotypes. This underscores the importance of incorporating time-series clinical data during ICU stays for a nuanced understanding of TBI phenotype progression.

Table 4. Key demographics and non-temporal clinical features of TBI patients in each phenotype.

| Feature | TRACK-TBI | | | MIMIC-IV | | |
|---|---|---|---|---|---|---|
| | Phenotype α | Phenotype β | Phenotype γ | Phenotype α | Phenotype β | Phenotype γ |
| Total subjects, n | 1557 | 320 | 283 | 1710 | 801 | 629 |
| Age | | | | | | |
| Age, mean ± SD | 38 ± 18 | 40 ± 19 | 43± 18 | 59 ± 22 | 61 ± 22 | 69 ± 20 |
| Age≤30, n (%) | 671 (43%) | 109 (34%) | 94 (33%) | 274 (16%) | 125 (16%) | 44 (7%) |
| 30<Age≤45, n (%) | 348 (23%) | 66 (21%) | 67 (24%) | 222 (13%) | 75 (9%) | 49 (8%) |
| 45<Age≤60, n (%) | 300 (19%) | 80 (25%) | 75 (26%) | 347 (20%) | 144 (18%) | 103 (16%) |
| Age>60, n (%) | 238 (15%) | 65 (20%) | 47 (17%) | 867 (51%) | 457 (57%) | 433 (69%) |
| Sex | | | | | | |
| Male, n (%) | 1059 (68%) | 237 (74%) | 216 (76%) | 1072 (63%) | 507 (63%) | 391 (62%) |
| Female, n (%) | 498 (32%) | 83 (26%) | 87 (24%) | 638 (37%) | 294 (37%) | 238 (38%) |
| Clinical variables | | | | | | |
| ED Glucose, mean ± SD | 130 ± 50 | 171 ± 65 | 150 ± 60 | 133 ± 53 | 139 ± 39 | 135 ± 37 |
| ED Hemoglobin, mean ± SD | 14.3 ± 1.6 | 13.3± 1.8 | 14 ± 1.7 | 12.5 ± 1.7 | 11.4 ± 1.9 | 12.2 ± 1.6 |
| ED INR, mean ± SD | 1.07 ± 0.2 | 1.24 ± 0.8 | 1.10 ± 0.2 | 1.25 ± 0.1 | 1.30 ± 0.1 | 1.26 ± 0.1 |
| ED White Blood Cell, mean ± SD | 11.6 ± 4.7 | 15.5 ± 6.4 | 13.7 ± 6.6 | 11.7 ± 5.7 | 12.9 ± 7.7 | 11.9 ± 4.7 |

4.2. Comparison of Outcomes of TBI Phenotypes

In the evaluation of identified TBI phenotypes within the TRACK-TBI and MIMIC-IV datasets, the study focuses on mortality rates and lengths of ICU stays as outcome endpoints. These outcomes were selected for their consistent availability across both datasets.

*Mortality Rate:* The overall mortality rates show distinct differences between the datasets, with MIMIC-IV exhibiting a mortality rate of 12%, in comparison to 6% in TRACK-TBI. Despite this variation, the stratification by phenotype shows a consistency. Specifically, phenotype β exhibits significantly higher mortality rates within both datasets: 33.4% in TRACK-TBI and 41.2% in MIMIC-IV. This increased rate in MIMIC-IV may be explained by the higher average age of phenotype β within this dataset, potentially reflecting an age-related increase in susceptibility to TBI complications. On the other hand, phenotypes α and γ maintain closely comparable mortality rates across both datasets. Phenotype α records a mortality rate of 1.3% in TRACK-TBI and 1.5% in MIMIC-IV, while phenotype γ shows rates of 3.9% and 4.9%, respectively.

*Length of ICU Stay:* The average ICU stay duration across datasets supports the consistency of TBI phenotype in two datasets (Figure 2). For phenotype α, the difference is minor, with a 5-day ICU stay in TRACK-TBI compared to 6 days in MIMIC-IV. Phenotype γ shows similar durations, averaging a 10-day stay in both datasets. However, phenotype β reveals a greater discrepancy—a 25-day average stay in TRACK-TBI versus a 15-day stay in MIMIC-IV. Several factors might explain this difference: the higher mortality rate for phenotype β in MIMIC-IV could lead to shorter ICU stays due to earlier patient deaths. Additionally, differences in clinical protocols, interventions, and resources specific to each dataset's source might influence recovery paths or treatment results.

Examining these outcomes, it is evident that TBI phenotypes in both datasets share strong similarities, emphasizing their potential for generalization.

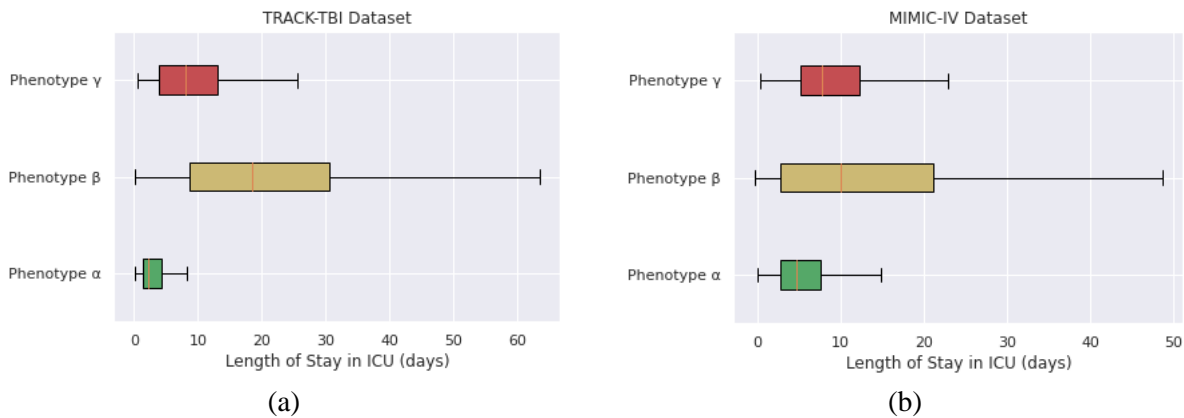

Figure 2. Boxplot of ICU stay duration for TBI phenotypes in TRACK-TBI (a) and MIMIC-IV (b).

4.3. Analysis of Time-Series Clinical Markers for TBI Phenotypes

*Level of Consciousness*: The GCS score, comprising eye, verbal, and motor components, is a measure for assessing the level of consciousness in TBI patients. Of these components, studies emphasize the eye and motor responses as particularly predictive of the TBI GOSE score [30,31]. Although there are significant differences in the GCS eye and motor scores of TBI phenotypes within each dataset ($p < 0.05$), both datasets present comparable consciousness patterns (Figure 3 and 4). Phenotype β is characterized by its severe neurological deficits, with the majority of these patients scoring the lowest possible GCS of 1 for both eye and motor responses, indicating no observable response. This profound impairment persists across the initial days of intensive care, marking the phenotype with a particularly severe clinical trajectory. In contrast, phenotype α is associated with relatively mild neurological deficits. Most patients in this phenotype achieve a GCS eye score of 4, denoting spontaneous eye-opening, and a motor score of 6, demonstrating the ability to follow commands—signs of substantially preserved consciousness and neurological function. The consciousness level of phenotype γ is intermediate between those of phenotypes α and β across both datasets. However, it is worth mentioning that phenotype γ is not equidistant between phenotypes α and β. Instead, its level of consciousness leans closer to that of phenotype α. This indicates that patients categorized within phenotype γ, despite exhibiting neurological challenges, generally experience less severe deficits compared to the profound impairments seen in phenotype β.

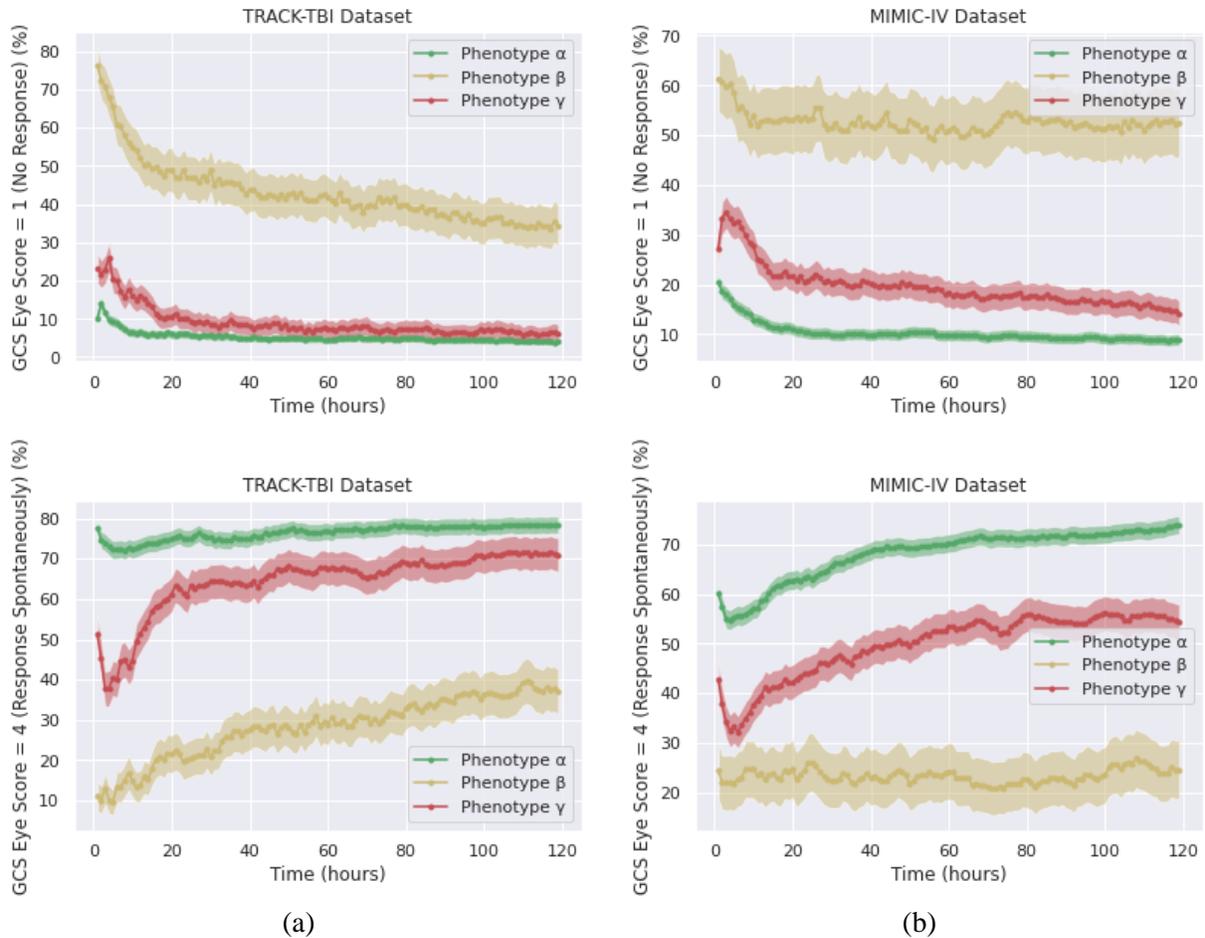

Figure 3. Percentage of the TBI patients with the lowest and highest GCS eye scores in TRACK-TBI (a) and MIMIC-IV (b) over the initial 120 hours in ICU.

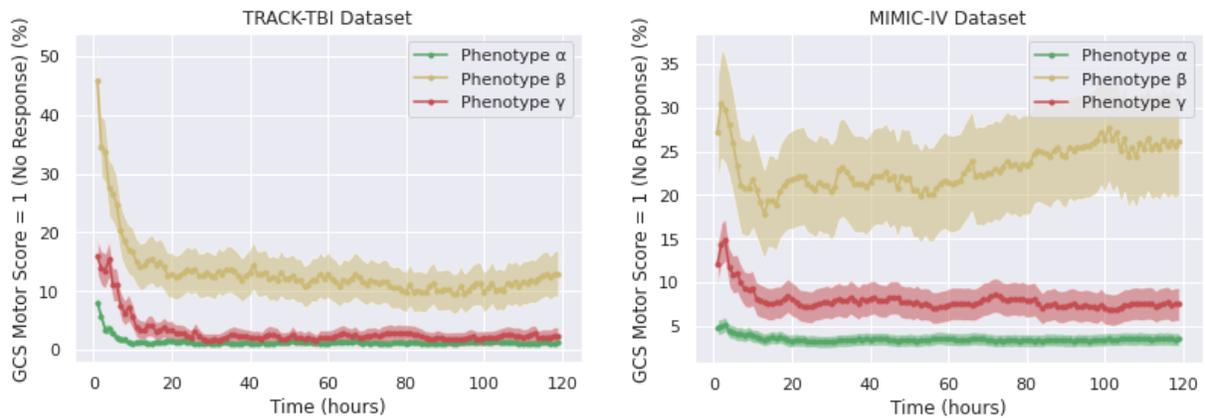

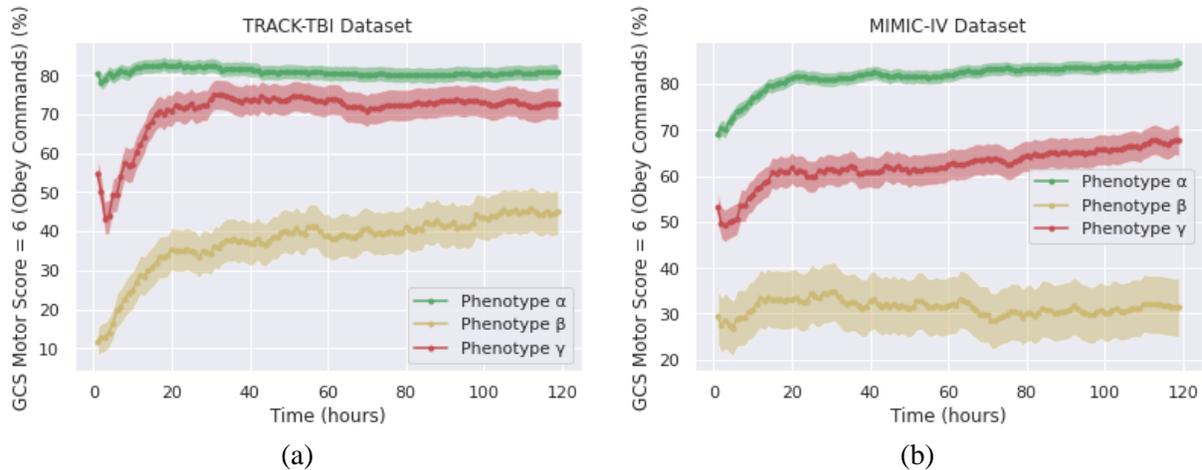

(a)                  (b)

Figure 4. Percentage of the TBI patients with the lowest and highest GCS motor scores in TRACK-TBI (a) and MIMIC-IV (b) over the initial 120 hours in ICU.

*Glucose Levels*: Figure 5 highlights distinct temporal trends of glucose levels among TBI phenotypes during early ICU admission across both datasets ($p < 0.05$). The patterns observed in TRACK-TBI and MIMIC-IV for TBI phenotypes are similar, underscoring the consistency of TBI phenotypes across these datasets. A notable trend is the initial steep decrease in glucose levels for all phenotypes, potentially indicating a stress-induced hyperglycemic response, which stabilizes subsequently. Phenotype β consistently displays elevated glucose levels, suggesting heightened physiological stress or pronounced metabolic dysfunction, often associated with severe TBI cases. In contrast, phenotypes α and γ exhibit moderate glucose levels, with γ slightly higher than α.

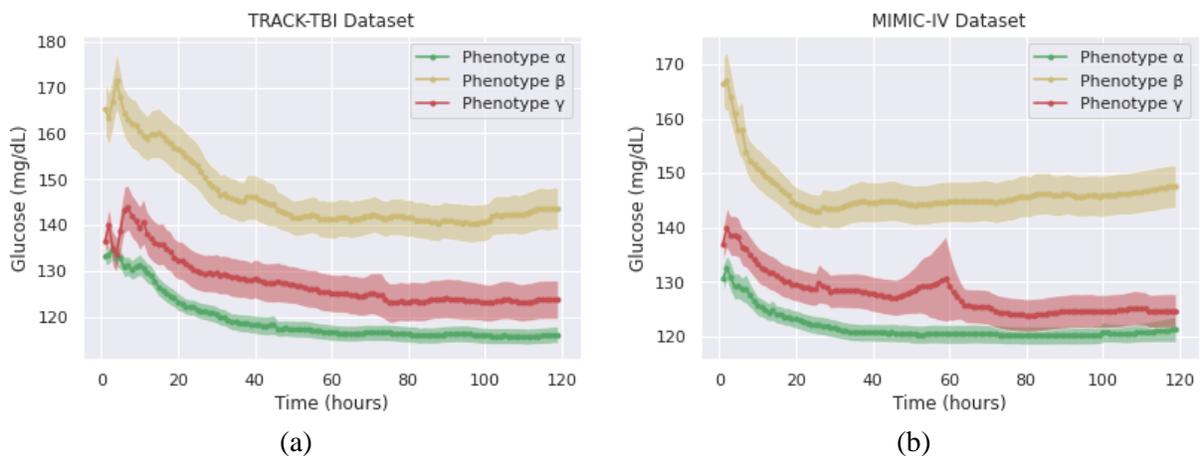

(a)                  (b)

Figure 5. Average glucose levels of TBI phenotypes in TRACK-TBI (a) and MIMIC-IV (b) during the initial 120 hours in ICU.

*Hematocrit Levels*: Hematocrit levels, indicative of the proportion of red blood cells in blood, provide insights into potential hemorrhagic conditions or anemia [32,33]. There are significant differences between the hematocrit levels of TBI phenotypes in each dataset ($p < 0.05$). Figure 6 reveals pronounced hematocrit levels in phenotype α, suggesting conditions such as dehydration or reduced plasma volume. Conversely, phenotype β displays the lowest hematocrit levels, hinting at potential hemorrhagic complications or dilutional anemia. A consistent pattern observed across all phenotypes is an initial decline in hematocrit levels, followed by stabilization. This trend may reflect a common post-injury physiological response, influenced by factors like hemorrhage, fluid resuscitation, or systemic inflammation, underscoring the importance of hematocrit as a dynamic marker in TBI management and prognosis.

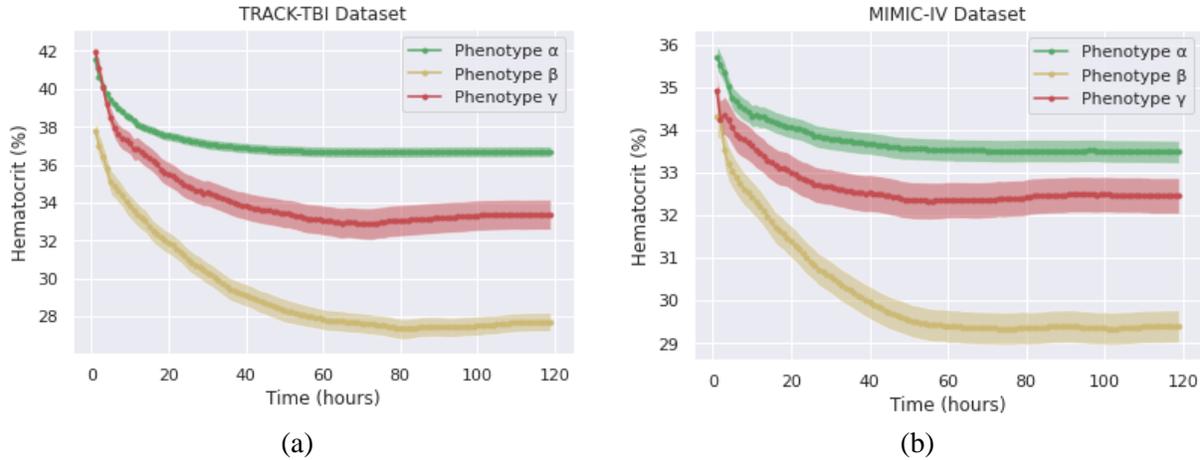

(a)  (b)

Figure 6. Average hematocrit levels of TBI phenotypes in TRACK-TBI (a) and MIMIC-IV (b) during the initial 120 hours in ICU.

*Hemoglobin Concentrations*: Hemoglobin, a key marker for evaluating oxygenation [34], displays significantly different trends across TBI phenotypes in each dataset ($p < 0.05$). Moreover, the TBI phenotypes in the TRACK-TBI and MIMIC-IV datasets exhibit similar patterns, emphasizing the consistency in hemoglobin trends of TBI phenotypes across these datasets. Phenotype α consistently exhibits the highest hemoglobin concentrations, suggesting optimal oxygen-carrying capacity (Figure 7). In contrast, phenotype β has the lowest hemoglobin levels, indicating potential challenges such as anemia or hemorrhagic events. All phenotypes initially show a decline in hemoglobin concentrations, which then level out. This pattern might stem from an immediate hemorrhagic response after TBI, compounded by hemodilution (dilution of blood) from medical fluid resuscitation. Additionally, trauma-induced systemic inflammation, which involves a body-wide response that can shift fluids, further influences hemoglobin concentrations. The subsequent stabilization of hemoglobin levels indicates effective early medical interventions and the body's adaptive responses.

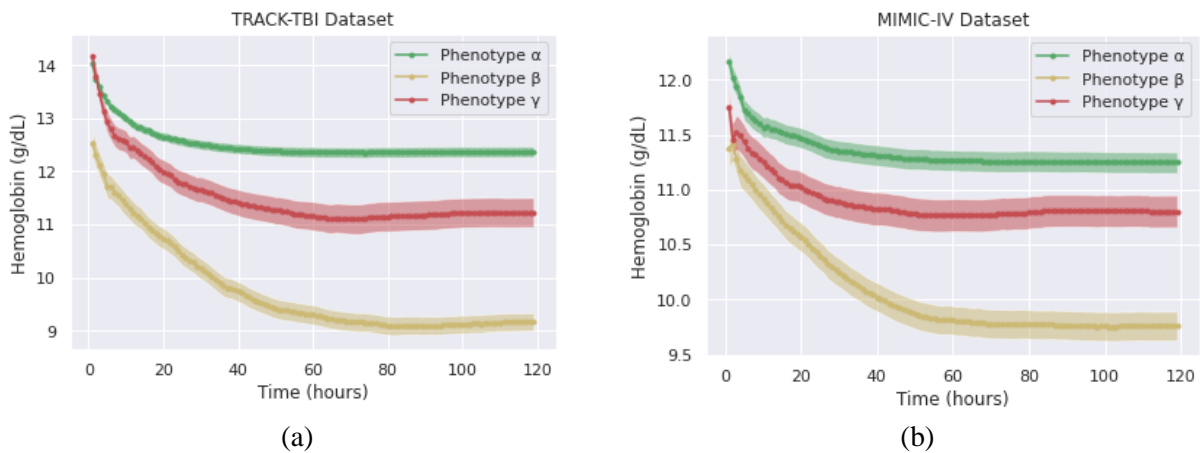

(a)  (b)

Figure 7. Average hemoglobin levels of TBI phenotypes in TRACK-TBI (a) and MIMIC-IV (b) during the initial 120 hours in ICU.

*White Blood Cells (WBC)*: The WBC count serves as an indicator of the body's inflammatory response and offers valuable insights into the identified TBI phenotypes [35]. There is a significant difference between the TBI phenotypes within each dataset regarding the temporal values of WBC ($p < 0.05$). However, the corresponding TBI phenotypes between TRACK-TBI and MIMIC-IV display similar patterns throughout the ICU stay (Figure 8). Phenotype β consistently shows the highest WBC counts in both datasets, suggesting an intensified inflammatory response due to severe injuries or significant blood loss. In contrast, phenotype α exhibits the lowest WBC counts, indicating a milder inflammatory response. In both datasets, the WBC counts of TBI phenotypes decrease over the ICU stay duration. This trend underscores the body's ability to manage and gradually resolve the initial inflammatory response, possibly due to medical interventions, natural healing processes, or a combination of both.

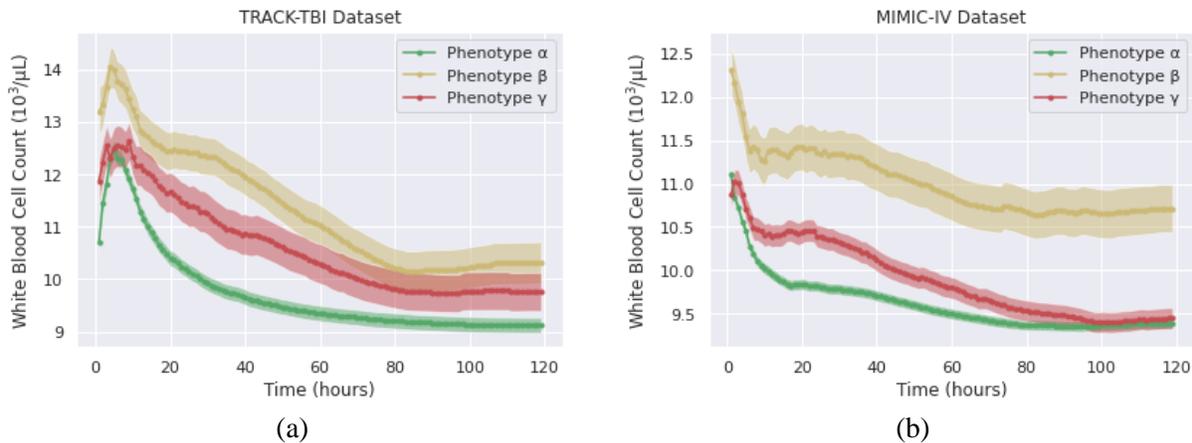

(a) (b)

Figure 8. Average hemoglobin levels of TBI phenotypes in TRACK-TBI (a) and MIMIC-IV (b) during the initial 120 hours in ICU.

*Coagulation Efficiency*: The ability to form clots and halt excessive bleeding is vital in managing traumatic injuries, including TBI. Key assessments for this include the INR and aPTT tests. Elevated INR values indicate a compromised clotting ability, increasing the risk of bleeding, while a prolonged aPTT suggests potential bleeding complications [36]. Our study identified distinct clotting patterns among TBI phenotypes within each dataset ($p < 0.05$). However, both TRACK-TBI and MIMIC-IV datasets demonstrated similar trends in terms of INR and aPTT values. Phenotype β consistently presented with elevated INR and aPTT levels (Figure 9). These patterns may indicate that patients with this phenotype are more prone to bleeding disorders, warranting close monitoring and potential interventions to manage bleeding risks. Conversely, phenotypes α and γ appear to maintain a more balanced coagulation profile, closer to standard reference ranges. Among these, phenotype α is characterized by the lowest INR and aPTT levels, suggesting a faster clotting ability and potentially a lower bleeding risk compared to other phenotypes. Notably, all three phenotypes in both datasets exhibited a rising trend in aPTT levels, possibly indicating evolving clotting dynamics due to TBI progression or treatment responses.

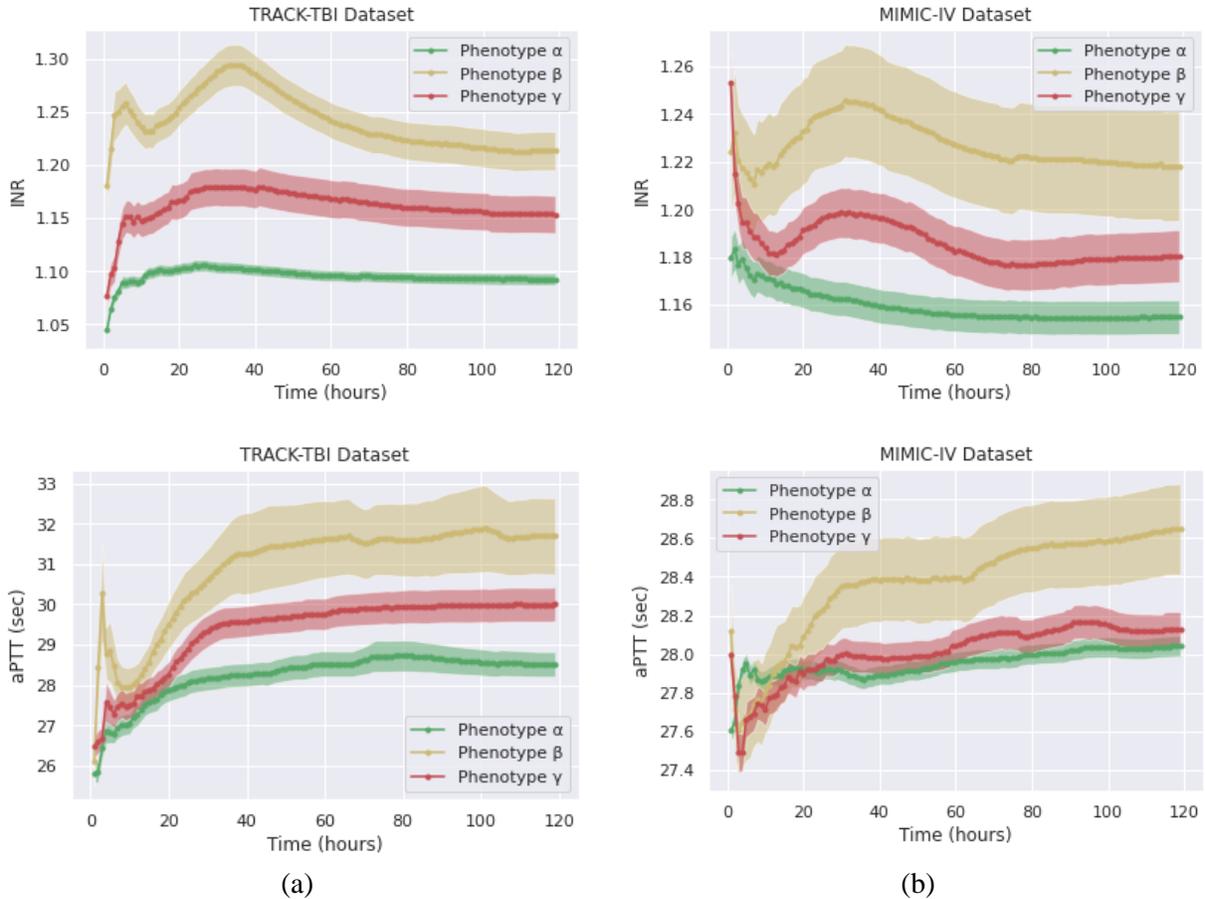

Figure 9. Average INR and aPTT levels of TBI phenotypes in TRACK-TBI (a) and MIMIC-IV (b) during the initial 120 hours in ICU.

## 4.4. Comparison of Heterogeneity and Dispersion across TBI Phenotypes

In this section, we used Principal Component Analysis (PCA) to identify the unique dispersion patterns of each TBI phenotype. Figure 10 presents the PCA plots, illustrating the clustering of TBI patients in the TRACK-TBI and MIMIC-IV datasets. Unlike our previous study on identifying TBI phenotypes [5], the current study leveraged learned representations rather than the raw input data for generating the PCA plots. This revealed distinct dispersion patterns for each phenotype and similar patterns across both datasets. In both datasets, phenotype α is characterized by a dense clustering of TBI patients. This compactness indicates a high degree of homogeneity within this group, suggesting a consistent set of clinical presentations and potentially a more predictable trajectory of progression and response to therapeutic interventions. On the other hand, phenotype β exhibits a more dispersed pattern in both datasets, representing its inherent variability. This spread signifies the diversity of clinical presentations and the complexities stemming from its severe TBI nature. This wide dispersion underscores the challenges in classifying and treating patients within this phenotype due to its diverse manifestations. Phenotype γ demonstrates a dispersion level that lies between the other two phenotypes in both datasets.

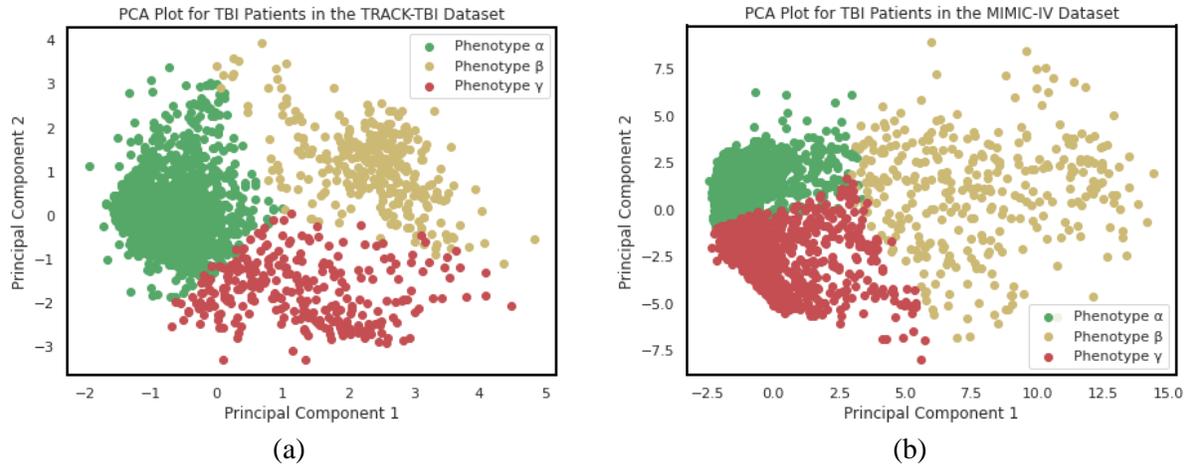

Figure 10. PCA visualization of TBI phenotypes from TRACK-TBI (a) and MIMIC-IV (b) using two principal components

4.5. Evaluation of Reproducibility of TBI Phenotypes

The classifier trained on the TRACK-TBI data demonstrated an average accuracy of 97%. Upon applying to the MIMIC-IV dataset, findings showed no significant difference in the distribution of patient phenotypes between the TRACK-TBI and the reproduced phenotypes in MIMIC-IV ($p > 0.05$). Similarly, a classifier trained on the MIMIC-IV dataset achieved an average accuracy of 95%. Upon application to the TRACK-TBI dataset, the analysis revealed no significant differences between the phenotypes identified in the MIMIC-IV dataset and the corresponding classifications in the TRACK-TBI dataset ($p > 0.05$). These findings underscore the reproducibility of the identified TBI phenotypes across external datasets, highlighting their potential for broad applicability and generalization.

## 5. Discussion

This study represents an application of the SLAC-Time clustering approach to discover generalizable TBI phenotypes from two distinct datasets: TRACK-TBI and MIMIC-IV. In the following sections, we delve deeper into the implications of our findings.

5.1. Robustness and Generalizability of the SLAC-Time Clustering Approach

The SLAC-Time clustering approach consistently exhibits optimal performance across both the TRACK-TBI and MIMIC-IV datasets when using identical hyperparameters, highlighting its reliability. The model's capacity to identify three clusters for TBI phenotypes across datasets suggests that the diverse manifestations of TBI may be condensed into three primary phenotypes. This finding holds significant clinical relevance, as it might simplify diagnostic procedures, treatment strategies, and prognosis assessments by classifying TBI patients into one of these three phenotypes.

5.2. Demographic Distributions: Age and Sex

There was a noticeable difference in the age demographics between the two datasets, with MIMIC-IV having an older population. This discrepancy in age demographics has potential clinical implications. Older populations often present with a wider range of comorbidities and may react differently to interventions than younger groups. The consistent age associations of phenotypes across datasets, where phenotype α represents the youngest and phenotype γ the oldest, may indicate underlying biological or physiological mechanisms that warrant further investigation.

Regarding sex distribution, both datasets showed a male predominance for TBI phenotypes. This inclination toward male TBI patients might reflect environmental and behavioral influences, suggesting that men may be more frequently exposed to environments or activities with a higher risk of TBI. This observation underscores the need for prevention strategies that focus on male-centric environments or activities.

5.3. Clinical Indicators and Outcomes

The consistent patterns observed across datasets indicate that phenotype α generally presents with milder clinical indicators, while phenotype β displays more pronounced physiological disturbances. These patterns have important clinical implications. Rapid and precise identification of these phenotypes could greatly influence and expedite the commencement of suitable treatments. The observed mortality rates underscore the urgency, particularly for phenotype β, which showed elevated rates. Although the older age demographic in MIMIC-IV may contribute to this observation, the inherent severity of phenotype β is evident and necessitates proactive clinical intervention. Furthermore, factors influencing the duration of ICU stays—whether they relate to mortality rates, clinical protocols, or specific patient attributes—offer valuable insights for health systems regarding resource distribution and patient care strategies.

5.4. Time-Series Clinical Markers

The GCS, glucose, hematocrit, hemoglobin, and WBC count provided rich insights into the dynamic progression of TBI manifestations across the early days of ICU admission. Such markers can aid clinicians in monitoring patient trajectories and adjusting interventions in real-time. The disparities in GCS scores among phenotypes emphasize the variation in neurological deficits faced by TBI patients. The marked neurological impairments in phenotype β, as indicated by their GCS scores, highlight the acute severity of this phenotype. This observation suggests that interventions tailored for phenotype β should prioritize addressing significant neurological dysfunctions. The relatively milder neurological disturbances in phenotypes α and γ necessitate different intervention strategies. Importantly, the consciousness levels observed in phenotype γ being closer to phenotype α rather than β is a clinically significant observation. It suggests that interventions beneficial for phenotype α might also be applicable, at least in part, to phenotype γ, but this requires further exploration.

Elevated glucose levels in phenotype β, potentially indicative of a stress-induced hyperglycemic response, emphasize the need for close monitoring and management. The consistency in hematocrit and hemoglobin trends across datasets underscores their value as TBI management markers. These trends, combined with the WBC count data, offer insights into the balance between hemorrhagic conditions and inflammation in TBI patients. Coagulation efficiency observations, particularly the clotting challenges in phenotype β, stress the importance of preemptive interventions to manage bleeding disorders. The rising aPTT trend across all phenotypes necessitates continued clinical vigilance.

5.5. Validation and Reproducibility

The correspondence between the distribution of TBI phenotypes and their reproduced counterparts in external datasets underscores the potential generalizability and robustness of the identified TBI phenotypes. This consistent reproducibility reinforces the reliability of the phenotypes for clinicians and strengthens confidence in their clinical relevance.

5.6. Limitations

Our study offers novel insights into generalizable TBI phenotypes. However, it is important to acknowledge several limitations. First, the scope of our research was limited by the range of clinical variables available. A more comprehensive analysis of phenotypic differences may be possible with inclusion of additional, detailed clinical variables. Specifically, incorporating data from imaging, detailed neurological symptoms, and multi-omics data (e.g., transcriptomics, proteomics, and metabolomics) could enrich the granularity of our phenotype clustering. In addition, the diversity within our datasets warrants critical examination. Despite their comprehensiveness, these datasets may not fully represent the global diversity of TBI patients. Factors such as ethnicity, geographic location, and socioeconomic status can influence the presentation and outcomes of TBI, yet these elements may not be adequately captured in our data. Furthermore, our study's focus on acute clinical markers does not encompass significant medical histories or chronic conditions of the patients. Other factors, such as post-injury rehabilitation practices, were also outside the scope of this research.

**6. Conclusions**

In our study, we employed the SLAC-Time clustering approach on multivariate time-series data from TRACK-TBI and MIMIC-IV datasets to identify generalizable TBI phenotypes. Our findings delineate three phenotypes: α, β, and γ, each with distinct characteristics but consistently present across both datasets. Phenotype α represents a group with favorable outcomes, characterized by the lowest mortality, shorter ICU stays, and predominantly younger patients. These patients exhibit good neurological outcomes, moderate metabolic stress, and show effective hydration, oxygenation, and coagulation profiles, coupled with a mild inflammatory response. Conversely, phenotype β encompasses patients with the most challenging clinical scenarios, indicated by the highest mortality, prolonged ICU stays, and severe neurological impairments. This group experiences significant metabolic stress, poor hydration and oxygenation, a strong inflammatory response, and a compromised coagulation profile. Phenotype γ, predominantly consisting of older patients, falls in between the other two, displaying moderate clinical severity. This group's characteristics are more aligned with phenotype α, demonstrating moderately elevated metabolic stress and more balanced hydration, oxygenation, and coagulation profiles. Notably, all phenotypes exhibit a male predominance, implying potential environmental or behavioral factors affecting TBI incidence. The consistent performance of SLAC-Time across datasets, with identical hyperparameters, underscores its robustness. The identification of three primary clusters indicates a potential simplification of TBI's inherent complexity. Supervised learning tests confirm the reproducibility of these phenotypes across datasets, suggesting their potential applicability to a broader TBI population. The findings of this study can serve as a foundation for the creation of precision interventions and predictive models tailored to these phenotypes, aiming to enhance TBI diagnosis and treatment. Future research could consider implementing explainability techniques to enhance the interpretation and transparency of the clustering results. Additionally, incorporating more detailed clinical variables, such as imaging, neurological evaluations, and multi-omics data, may help capture more subtle distinctions within phenotypes. Extensive studies across diverse demographics are needed for deeper validation and understanding. Furthermore, since there are no definitive ground truth labels for TBI phenotypes yet, future work may also focus on developing innovative phenotype validation methods.

## Acknowledgments

This material is based upon work supported by the National Science Foundation under grants #1838730 and #1838745. Dr. Foreman was supported by the National Institute of Neurological Disorders and Stroke of the National Institutes of Health (K23NS101123). The content is solely the responsibility of the authors. Any opinions, findings, and conclusions or recommendations expressed in this material are those of the authors and do not necessarily reflect the views of the National Science Foundation or of the National Institutes of Health. The authors acknowledge the TRACK-TBI Study Investigators for providing access to data used in this work.